\documentclass[10pt,twocolumn,letterpaper]{article}

\usepackage[pagenumbers]{cvpr} 

\usepackage[dvipsnames]{xcolor}

\definecolor{cvprblue}{rgb}{0.21,0.49,0.74}
\usepackage[pagebackref,breaklinks,colorlinks,citecolor=cvprblue]{hyperref}
\usepackage{booktabs}
\usepackage{siunitx}
\usepackage{multirow}
\usepackage{pifont}
\usepackage{xspace}
\usepackage{tabulary}
\usepackage{bm}
\usepackage{array}
\usepackage{placeins}
\usepackage{graphicx}

\usepackage{find2d_macros}

\def\paperID{127} 
\def\confName{3DV\xspace}
\def\confYear{2025\xspace}

\title{FOCUS - Multi-View Foot Reconstruction From Synthetically Trained Dense Correspondences}

\author{\mbox{Oliver Boyne \qquad Roberto Cipolla}\\
Department of Engineering, University of Cambridge, U.K.\\
{\tt\small \{ob312, rc10001\}@cam.ac.uk}
\and
}

\begin{document}
\maketitle
\begin{abstract}

Surface reconstruction from multiple, calibrated images is a challenging task - often requiring a large number of collected images with significant overlap. We look at the specific case of human foot reconstruction. As with previous successful foot reconstruction work, we seek to extract rich per-pixel geometry cues from multi-view RGB images, and fuse these into a final 3D object. Our method, FOCUS, tackles this problem with 3 main contributions: (i) \ourSynth, an extension of an existing synthetic foot dataset to include a new data type: dense correspondence with the parameterized foot model FIND; (ii) an uncertainty-aware dense correspondence predictor trained on our synthetic dataset; (iii) two methods for reconstructing a 3D surface from dense correspondence predictions: one inspired by Structure-from-Motion, and one optimization-based using the FIND model. We show that our reconstruction achieves state-of-the-art reconstruction quality in a few-view setting, performing comparably to state-of-the-art when many views are available, runs substantially faster, and can run without a GPU. We release our synthetic dataset to the research community. Code is available at: \url{https://github.com/OllieBoyne/FOCUS}

\end{abstract}    
\section{Introduction}
\label{sec:intro}

\begin{figure}
    \centering
    \includegraphics[width=\linewidth]{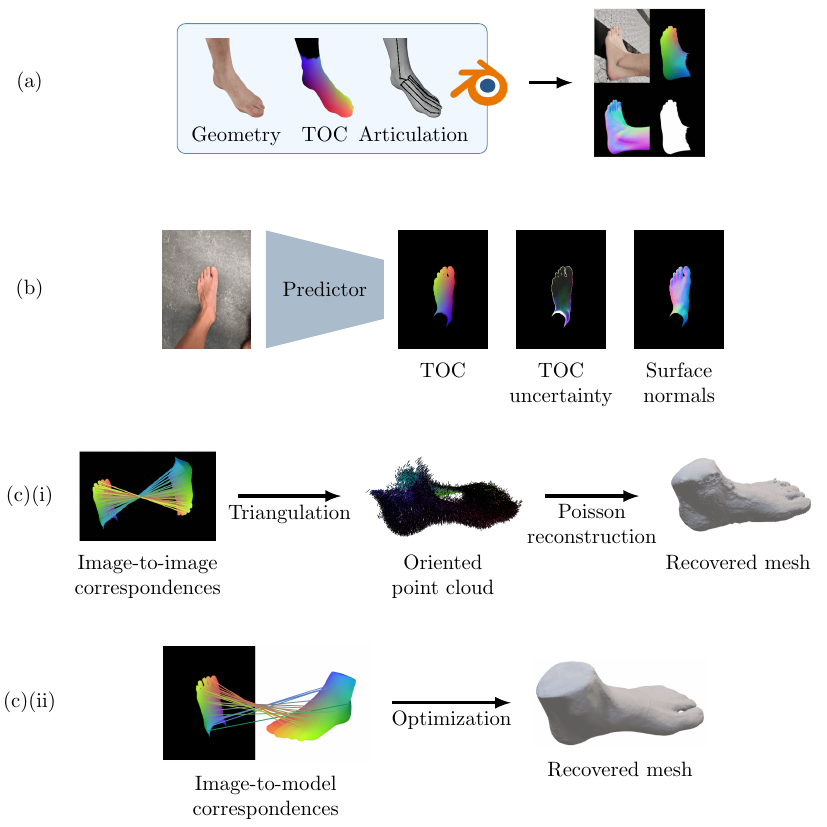}
    \caption{\textbf{Method overview.} (a) We use Blender \cite{blender} to render articulated high resolution meshes, with dense correspondences (TOC) to the generative FIND \cite{boyne2022find} model. (b) We train a model to predict TOCs and surface normals on real images. (c) We combine these predictions together in a multi-view setting via two methods to yield accurate surface reconstructions: (i) \ourSfM, a Structure-from-Motion based approach; and (ii) \ourOptim, a model fitting, optimization-based approach.}
    \label{fig:splash}
\end{figure}

Accurate 3D reconstruction of human body parts from images is a challenging computer vision task, of significant interest to the health, fashion and fitness industry. In this paper, we address the problem of reconstructing a human foot accurately from multiple views. Shoe retail, orthotics and health monitoring can all be improved with accurate models of the foot, and the growing digital markets for these applications has created a demand for recovering such models from mobile phone images captured by ordinary users.

Existing approaches to foot reconstruction include: (i) specialized scanning equipment \cite{ArtecLeo, Volumental}; (ii) reconstruction of unoriented point clouds from depth maps or phone-based sensors \cite{xesto, lunscher2017point}; (iii) photogrammetry pipelines, such as COLMAP, of Structure-from-Motion (SfM) followed by Multi-View Stereo (MVS) \cite{schonberger2016structure, schonberger2016pixelwise}; and (iv) fitting generative models to silhouettes, keypoints and surface normals \cite{kok2020footnet, boyne2024found}.

These methods come with substantial drawbacks: (i) expensive scanning equipment is not accessible to most consumers; (ii) phone-based sensors are limited in availability and ease of use, and noisy, unoriented point clouds are difficult to use for desired applications such as rendering and taking measurements; (iii) SfM and MVS require a large number of input views and favourable lighting conditions; and (iv) current optimization-based approaches use silhouettes, keypoints and surface normals, which only provide a limited representation of the full set of available cues in the image that can be used for accurate surface reconstruction.

To this end, we introduce FOCUS, \textbf{F}oot \textbf{O}ptimization via \textbf{C}orrespondences \textbf{U}sing \textbf{S}ynthetics - which improves on current multi-view reconstruction pipelines by predicting per-pixel correspondences relative to the FIND \cite{boyne2022find} parameterized foot model. We find this representation provides a much stronger geometric grounding than solely silhouettes \cite{kok2020footnet} and surface normals \cite{boyne2024found} seen in prior foot reconstruction work, allowing for a more direct training signal from which to reconstruct a foot model. We outline our method in Figure \ref{fig:splash}, and our key contributions are as follows:

\begin{itemize}
    \item In order to learn per-pixel correspondences, we extend prior research into foot synthetic data to release \ourSynth, a \textbf{large scale synthetic dataset} of 100,000 photorealistic foot images, coupled with accurate surface normals and dense correspondence labels. We drastically increase the background and lighting variety compared to \textit{SynFoot}, and introduce articulation into the scans to provide a greater pose variation.

    \item We introduce \textbf{Template Object Coordinates} (TOCs), normalized coordinates in the space of the FIND parameterized foot model, as a new representation of foot geometry. We train a predictor to jointly predict these per-pixel correspondences, in addition to a corresponding uncertainty. We follow a similar approach to prior work, to successfully train the network solely on synthetic data, and show plausible predictions on in-the-wild images.

    \item We introduce \ourSfM, a method for fusing multiple views of correspondence predictions into a single oriented point cloud, by matching correspondences and triangulating, similar to Structure-from-Motion. We use Poisson reconstruction \cite{kazhdan2013screened} to convert the oriented point cloud to a final mesh. This method does not require a GPU, requires fewer views than COLMAP, and is substantially faster than previous work.

    \item We also introduce \ourOptim, a method for directly optimizing the parameterized FIND \cite{boyne2022find} model to these dense correspondence images, producing a watertight, parameterized mesh as output. This method directly uses the predicted correspondence uncertainty, and is capable of accurate reconstruction on as few as 3 views.

\end{itemize}

\section{Related Work}

\paragraph{Synthetic dataset generation.} Synthetic rendering has become a growing source for data generation in the computer vision community. As photorealistic rendering capabilities have improved, synthetic pipelines have been used to produce high quality, large scale datasets, that are less expensive to scale than manually labelled datasets, and can often be more accurate. These pipelines are also useful for tasks that are difficult or impossible for human labellers, such as per-pixel labelling. Existing research has shown the viability of mostly, or entirely, synthetic data in training for complex downstream tasks - examples for human body reconstruction include bodies \cite{varol2017learning}, faces \cite{wood2021fake, bae2023digiface}, eyes \cite{wood2015rendering}, and feet \cite{boyne2024found}.

\paragraph{Multi-view reconstruction.} In order to recover geometry from multiple images, the relative camera positions and camera internal parameters must be known. These can be obtained directly from measurements of the image capturing device using onboard Inertial Measurement Units (IMUs), or can be recovered via sparse 3D reconstruction from Structure From Motion (SfM) \cite{schonberger2016structure}.

To recover the surface geometry, a common method is Multi-View Stereo (MVS) \cite{schonberger2016pixelwise}, a process of optimizing depth and normal maps across views, from which an oriented point cloud can be recovered. From this, a mesh is constructed using a surface reconstruction algorithm \cite{kazhdan2013screened}. COLMAP \cite{schonberger2016structure, schonberger2016pixelwise} is a popular implementation of this full pipeline.

In recent years, neural rendering for surface reconstruction has grown in interest \cite{mildenhall2021nerf, yariv2020multiview}. In these methods, a neural representation of a 3D scene is trained, which, when rendered through some differentiable process, accurately reconstructs reference views. These methods often require large amounts of training time and compute, and large numbers of input views for accurate reconstructions.

\paragraph{Dense correspondences.} 

For pose and shape reconstruction, sparse correspondences, or keypoints, are often used for registration. Obtaining accurate keypoint labels is fast and easy to explain to human labellers.

To fully reconstruct an accurate surface, dense correspondences can be incredibly useful - a mapping from every pixel in one image to a pixel in another.
Some implementations treat this as a generic, image pair matching problem - often called \textit{optical flow}. Finding this flow field for image pairs has been shown to be solvable via both optimization \cite{liu2010sift, taniai2016joint} using image features, or by training a model to directly predict this flow \cite{sun2018pwc, rocco2017convolutional}.

Other implementations seek to match all pixels of an object in an image to that object's local space, normalized to a unit cube. This approach, introduced by \citet{wang2019normalized} as Normalized Object Coordinates (NOCs), has been used to reconstruct object size and pose from single images, and by \citet{gumeli2023objectmatch} for multi-view joint object pose and camera registration.

For matching an object of a known category, it can be more advantageous to be able to map any number of images to some canonical object space - this allows for more effective fusion of multiple views, and for matching across a wider range of viewpoints.
Obtaining the data necessary to learn this canonical mapping can be difficult. For human body estimation, \citet{guler2018densepose} inferred dense correspondences from sparse human labelling and part matching. \citet{taylor2012vitruvian} directly learn dense correspondences from depth images via regression forests.

\citet{zeng20203d} learn to predict a mapping between pixels of an image of a human, and a UV map of a human body mesh, and task a regressor to directly recover a 3D surface from this prediction on a single image.

We also learn a mapping directly to a parameterized mesh. We do this entirely synthetically, and are able to leverage these correspondences in a multi-view setting to obtain an accurate reconstruction.

\paragraph{Human foot reconstruction.} Obtaining accurate models of the human foot is of significant interest to the footwear and orthotics industries. The prevalence of mobile phones and digital shopping and health has increased the demand for these reconstruction methods to be made possible with data collected from consumer devices.

Some methods collect point cloud data \cite{lunscher2017point, xesto}, which can be achieved with LiDAR or structured light sensors available on certain mobile phones. Such sensors are not universally available, and often these point clouds are noisy and not capable of providing a detailed, realistic foot surface.

Other methods instead seek to fit parameterized models of feet to input data. Earlier works built Principal Component Analysis (PCA) models of feet from sampling vertices from photogrammetry \cite{kok2020footnet} and scanners \cite{amstutz2008pca}. \citet{kok2020footnet} also fit their PCA model to predicted image silhouettes.

The PCA approach has been improved in recent years by \citet{osman2022supr}, with SUPR, a PCA model of the human foot to be combined with the SMPL \cite{SMPL:2015} full body model for the task of expressive, full body reconstruction. 

Another approach is the FIND model \cite{boyne2022find}, a non-linear generative model of the foot. Rather than PCA, FIND uses an implicit network to deform a template mesh per-vertex to a target pose and shape. With this model, Foot3D was released - a collection of high resolution foot scans.

Recently, FOUND \cite{boyne2024found} introduced a method of optimizing the parameters of the FIND model to match with predicted surface normals in 2D in a multi-view setting. Due to a lack of available real surface normal data, the surface normal predictor was trained in a synthetic setting, generalizing well to in-the-wild images.
\section{Method}

\subsection{Template Object Coordinates}

\begin{figure}
    \centering
    \includegraphics[width=\linewidth]{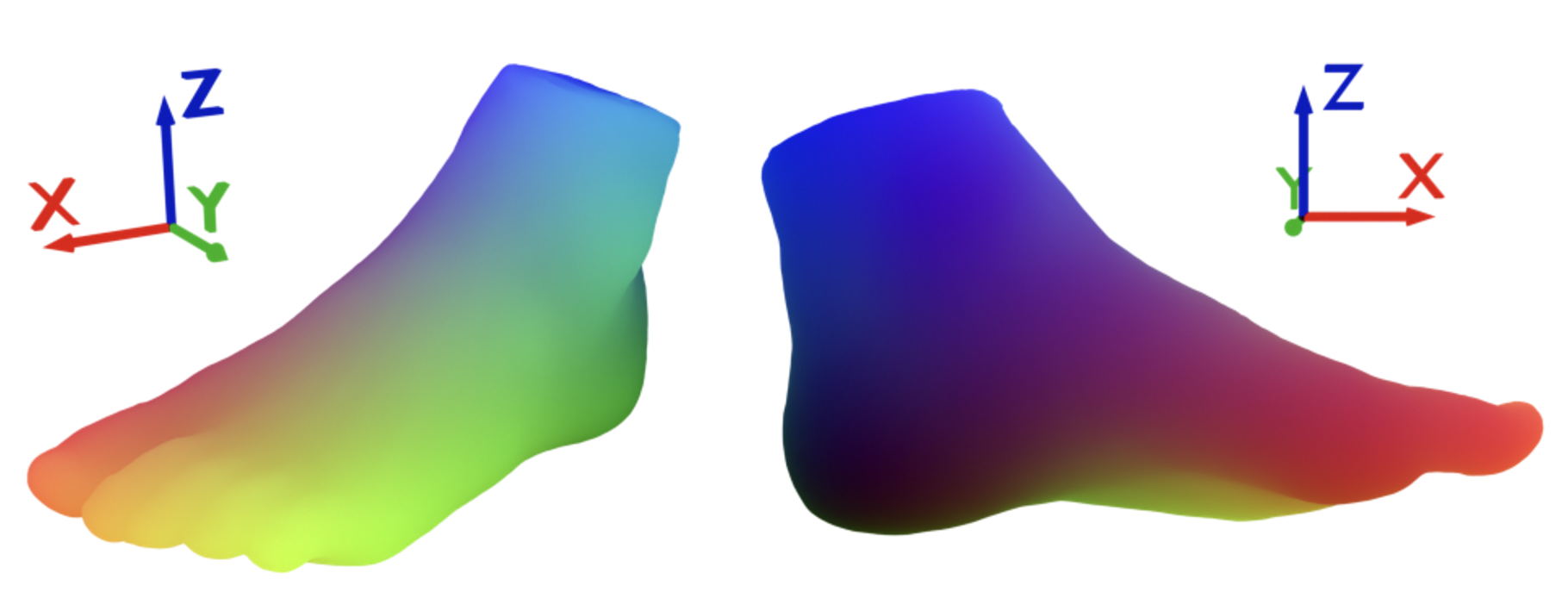}
    \caption{\textbf{TOC definition.} Template Object Coordinates (TOCs), shown on the template of the FIND mesh. RGB values correspond to XYZ, normalized to 0-1 within the template space.}
    \label{fig:toc_definition}
\end{figure}

We seek to define some dense correspondence over a foot surface.
Many dense correspondence predictors, \cite{guler2018densepose, taylor2012vitruvian, zeng20203d}, especially for human body prediction, use a 2D parameterization of the surface, such as a UV mapping. We choose to use a 3D parameterization, for three reasons: (i) ease of construction; (ii) a more intuitive uncertainty representation; and (iii) consistency with the coordinate representation used in the FIND \cite{boyne2022find} model, which is key to both our synthetic data and reconstruction approaches.

We therefore define \textbf{Template Object Coordinates} (TOCs), denoted as $\toc$. Visualized in Figure \ref{fig:toc_definition}, TOC values represent points in the 3D local space of the FIND model's template mesh, normalized to the template's bounding box.

Where Normalized Object Coordinates (NOCs) simply map to points in an object's local space, TOCs map points from any deformed, articulated foot directly to the template's local space. This is essential to directly fit the FIND model to TOC predictions, and allows for our per-pixel predictive network in Section \ref{sec:predictor} to predict correspondences agnostic to pose, shape and identity.

\subsection{\ourSynth}

We extend SynFoot \cite{boyne2024found}, a large scale synthetic dataset for feet: we add articulated feet, our new TOC representation, and increased background and lighting diversity.

\begin{figure}
    \centering
    \begin{tabular}{ccc}
    \includegraphics[width=0.3\linewidth]{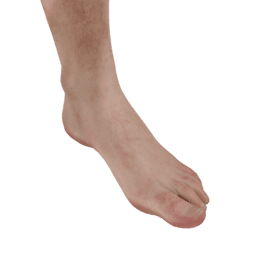} &
    \includegraphics[width=0.3\linewidth]{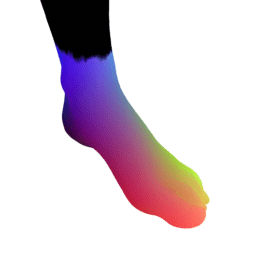} &
    \includegraphics[width=0.3\linewidth]{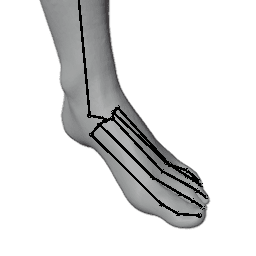} \\
    (a) & (b) & (c)
    \end{tabular}
    \vspace{-5pt}
    \caption{\textbf{Models for rendering.} One of 8 foot models used for the synthetic dataset. The mesh has (a) geometry and texture, (b) a TOC mapping to the FIND template model, and (c) a skeleton used for articulation.}
    \label{fig:0015-A}
\end{figure}

\paragraph{Articulation.} We manually add a skeleton to all 8 meshes in the original dataset, in line with anatomical diagrams \cite{houglum2011brunnstrom}. We use Blender's automatic weight calculation \cite{baran2007automatic} to handle the vertex weighting.

We identify 5 methods of articulation described in foot literature \cite{houglum2011brunnstrom}: dorsiflexion/plantarflexion (pitch), lateral/medial rotation (yaw), inversion/eversion (roll), toe extension/flexion (up/down), and toe abduction/adduction (outwards/inwards). We randomly sample these poses and apply combinations of them to our meshes to provide variation in articulation in the synthetic data.

\paragraph{TOCs.} We render TOCs for all meshes, by first fitting the FIND mesh to each model as in the original FIND paper \cite{boyne2022find}, and using this to find a vertex-to-vertex mapping between each mesh and the FIND template space. This provides us with per-vertex TOC values for each mesh to be used for our synthetic data, as in Figure \ref{fig:0015-A}.

\begin{figure}
    \centering

    \begin{tabular}{*{4}{@{}Q{0.25\linewidth}}@{}}
    (a) & (b) &  (c) & (d)
    \end{tabular}

    \includegraphics[width=\linewidth]{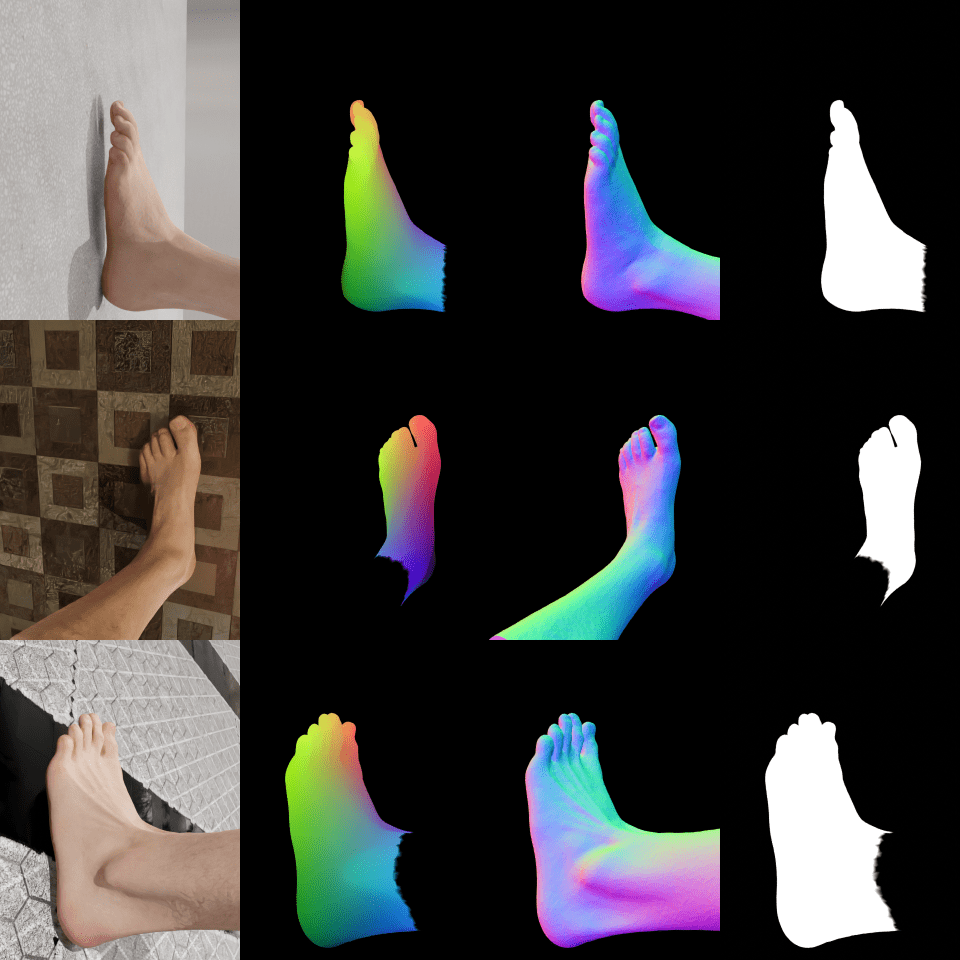}
    \caption{\textbf{\ourSynth examples.} We show (a) RGB, (b) TOC, (c) surface normals, and (d) segmentation masks. Further examples are included in the supplementary material.}
    \label{fig:synth_examples}
\end{figure}

\paragraph{Rendering.} We use the BlenderSynth \cite{boyne2024found} package to render the dataset. We drastically increase the quantity and variety of HDRIs and background textures compared to SynFoot \cite{boyne2024found}, increasing the number of HDRIs from 14 to 733, and background textures from 34 to 541 using Poly Haven assets \cite{polyhaven}. We add a new shader to render TOCs. We render 100,000 images at 480 x 640 resolution. Figure \ref{fig:synth_examples} shows some examples of our synthetic dataset.
\mynote{HDRIs need defining?}

\subsection{Training a predictor}
\label{sec:predictor}

\begin{figure}
    \begin{tabular}{*{3}{@{}Q{0.237\linewidth}}@{}}
    (a) & (b) & (c)
    \end{tabular}

    \centering
    \includegraphics[width=\linewidth]{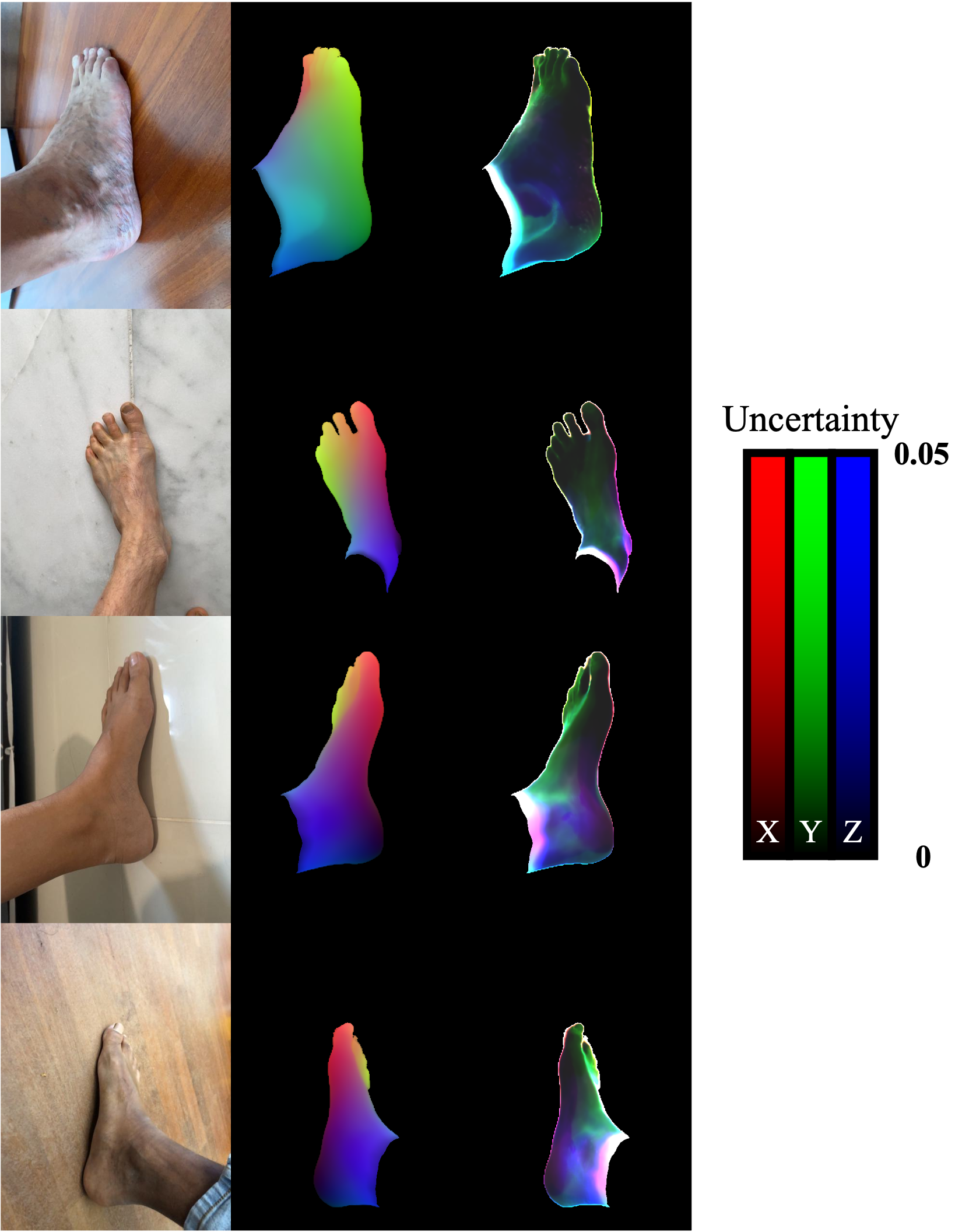}
        \caption{\textbf{TOC in-the-wild predictions.} Predictions on real images, showing (a) RGB input, (b) TOC $\toc$, (c) TOC uncertainty $\tocstd$. Further examples are included in the supplementary material.}
    \label{fig:toc_itw}
\end{figure}

\def\tunc{\tau}

We start with a model used to make coarse surface normal and uncertainty predictions, and uncertainty-guided pixel-wise refinements, as in \cite{alhashim2018high, bae2021estimating}. We modify the output prediction heads to tackle our downstream tasks.

First, we add a head to predict a binary segmentation heatmap for the foot using Binary Cross-Entropy loss. At inference time, we threshold all predictions according to this heatmap being larger than 0.5.

Next, we add heads to predict a probability distribution for the TOC values.
We model our TOC prediction as a normal distribution in XYZ,

\begin{equation}
    \toc \sim \mathcal{N}(\tocmean, \tocvar),
\end{equation}

\noindent and have one head to predict $\tocmean$, and another to predict $\log \tocvar$, independently in each axis XYZ. We combine these predictions in an uncertainty-aware loss during training of the predictor, which minimizes the negative log-likelihood of the TOC predictions,

\begin{equation}
    \loss{TOC} = \left\| \frac{\left( \tocmean - \toc_{\textrm{gt}}\right)^2}{\tocvar} \right\|_2 + \log{\tocvar}.
\end{equation}

We show examples of TOC inference on in-the-wild images in Figure \ref{fig:toc_itw}.

\subsection{\ourSfM}
\label{sec:point_cloud_fusion}

\begin{figure}
    \includegraphics[width=\linewidth]{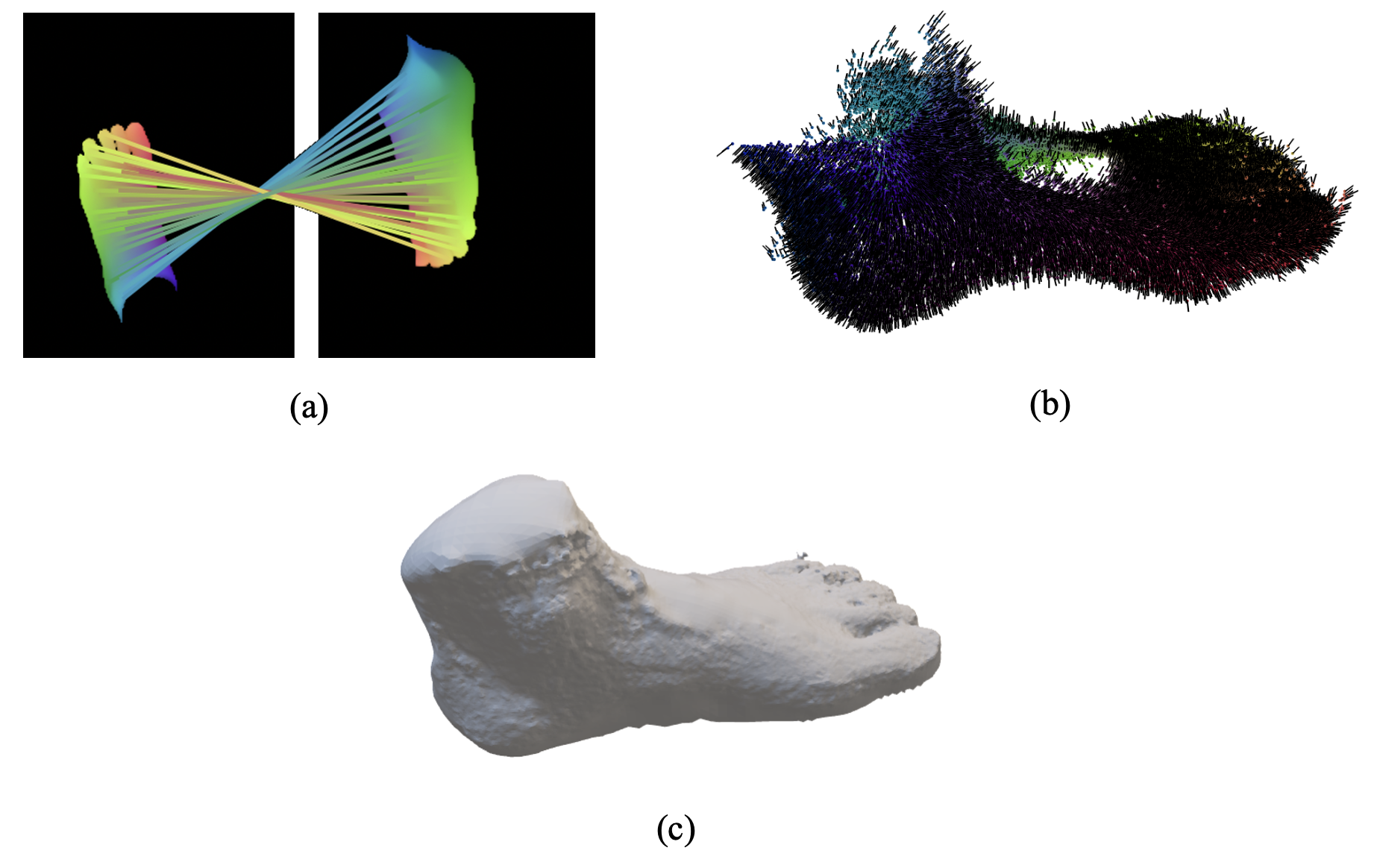}
    \caption{\textbf{\ourSfM overview.} (a) We find correspondences between images by matching TOC values; (b) we triangulate and collect normals across views to construct an oriented point cloud; (c) we use Poisson surface reconstruction to form a final mesh.}
    \label{fig:focussfm}
\end{figure}

Given these TOC predictions, we now wish to reconstruct a surface. The first of our two proposed methods for reconstruction, \ourSfM, is outlined in Figure \ref{fig:focussfm} and takes inspiration from traditional Structure-from-Motion.

We predict TOCs, segmentation masks and surface normals on $N$ images of a captured foot with known camera extrinsics and intrinsics. \ourSfM aims to identify correspondences between these predictions, and use triangulation to reconstruct an oriented point cloud.

\paragraph{Sampling. } For each image, we sample $P$ points within the mask of prediction. For each point, we collect the value of the pixel position, TOC, and surface normal.

\paragraph{Pixel-level correspondences. } We now want to find correspondences between images, where a `correspondence' refers to an exact TOC value. Each correspondence might only appear in a subset of images. We employ an efficient approach using nearest neighbor sampling. For each image, we structure all TOC values as a KD-tree \cite{bentley1975multidimensional} and, for each set of $P$ correspondences, look up the nearest neighbor in each image by $\ell2$ distance.

\paragraph{Subpixel correspondences. } We achieve more accurate correspondences by matching at the sub-pixel level. We use bilinear interpolation to upscale the 3x3 patch around each found correspondence by a factor of 8, and then search within this window for a better match. This search is implemented in Cython \cite{behnel2010cython} for performance.

Once we have the match, we capture the surface normal at each point. We only consider a correspondence found if the TOC value is within 0.002 $\ell 2$ distance of the original sampled value.

The result of this is $C \leq NP$ correspondences, each containing a subset of the $N$ views, and pixel positions for each view.

\paragraph{Triangulation. } For each correspondence, we triangulate across all views in which that correspondence is present, using the known camera parameters, and the Direct Linear Transformation \cite{hartley2003multiple}. This provides a point cloud of triangulated correspondences.

\paragraph{Filtering. } We filter the collected point cloud via three methods: (i) removing points whose average reprojection error is above a threshold; (ii) removing all points below the floor (Z=0); (iii) statistical outlier removal \cite{kriegel2009loop}.

\paragraph{Normal aggregation. } We also collect normals along the correspondences. To provide a normal estimate, we convert all normals to spherical coordinates $(1, \theta, \phi)$, and average over $\theta$ and $\phi$ to reach a consensus normal.

\paragraph{Poisson surface reconstruction. } Now that we have an oriented point cloud, we use Screened Poisson Reconstruction \cite{kazhdan2013screened} in Meshlab \cite{meshlab} to reconstruct a surface.

\paragraph{Implementation details.} The hyperparameters chosen in this process, such as for filtering and Poisson reconstruction, can be found in the supplementary material.

\subsection{\ourOptim}

The second approach we introduce, \ourOptim, takes inspiration from FOUND \cite{boyne2024found} - fitting a parameterized model directly to predictions made in image space.
We seek to optimize the global transformation ($r,s,t$) and FIND shape and pose embeddings ($z_s, z_p$).

Rather than use differentiable rendering, as in FOUND, the TOC representation allows us to sample points on the image, and optimize the FIND model such that the corresponding point on the FIND model projects onto the same pixel position. This is essentially a keypoint loss, where any number of keypoints can be sampled at arbitrary locations in image space.

\def\J{\mathbf{J}}
\def\predpixel{\bm{\hat\pixel}}

\paragraph{Sampling.} As in Section \ref{sec:point_cloud_fusion}, we sample $P$ $\toc$ values for each of the $N$ images within the mask of prediction, as well as recording their pixel position $\pixel$.

The TOC values $\toc$ are in normalized space - we convert them to FIND space, $\toc'$, by mapping to the axis aligned bounding box of the FIND template mesh.

\paragraph{FIND model.} As defined in \cite{boyne2022find}, the FIND model maps a 3D point on the surface of a template mesh $\mathbf{x_1}$ to a deformed mesh, under shape and pose embeddings ($z_s, z_p$), and global transformation  ($r,s,t$), to a deformed point $\mathbf{x_2}$,

\begin{equation}
    \mathbf{x_2} = F(\mathbf{x_1}, z_s, z_p, r, s, t)
\end{equation}

\paragraph{Projection.} Under our FIND and camera models, we want to project our $\toc'$ estimates onto the image plane. Given a function that maps a world point to pixel space under a given camera model, $f$, and the FIND model $F$, a reprojected point in 2D space can be calculated,

\begin{equation}
    \predpixel = f\left(F\left(\toc', z_s, z_p, r, s, t\right)\right).
\end{equation}

For our experiments, we use the default camera model of COLMAP \cite{schonberger2016structure} - a simple pinhole camera.

\paragraph{Uncertainty. } As well as predicting per-pixel TOC values, our predictor also provides per-pixel TOC uncertainties, $\bm{\sigma}_{\toc}$. We can use this to weight our reprojection prediction. To do this, we need to propagate the uncertainty to calculate an uncertainty in pixel position, $\bm{\sigma}_{\predpixel}$. We use auto-gradient computation in PyTorch \cite{paszke2017automatic} to calculate the Jacobian $\J$ of the transformation from $\toc$ to $\predpixel$.
Next, we transform the uncertainty using the first order approximation given by \citet{ochoa2006covariance},

\begin{equation}
    \bm{\Sigma_{\predpixel}} \approx \J \bm{\Sigma_{\toc}} \J^T, \quad \textrm{where} \, \bm{\Sigma_x} = \operatorname{diag}(\bm{\sigma}_x^2).
\end{equation}

\paragraph{Training loss.} We now have a predicted pixel position $\predpixel$, and associated uncertainty $\bm{\sigma}_{\predpixel}$. We construct a loss which minimizes the average projected pixel error, weighting samples according to their uncertainty,

\mynote{Check if std should be var}
\begin{equation}
    \loss{\textrm{\ourOptim}} = \frac{1}{NP} \sum^{NP}_{i=1}{\ltwo{\frac{\predpixel_i - \pixel_i}{\bm{\sigma}_{\predpixel_i}}}}.
\end{equation}

We minimize this loss by optimizing the parameters ${r, s, t, z_p, z_s}$. Similarly to FOUND \cite{boyne2024found}, we use a two-stage optimization process - first optimizing just registration parameters ${r,s,t}$, and then all parameters. Each stage runs for 500 epochs, using an Adam optimizer \cite{kingma2014adam} with a learning rate of 0.001.
\section{Experiments}

\paragraph{3D baseline. } We evaluate on Foot3D, a baseline 3D foot reconstruction dataset released in \cite{boyne2024found} - 14 scenes of real scanned feet, with a total of 474 calibrated images. All methods are provided with the same set of input images and camera calibrations.

As the task focuses on reconstruction of the foot (not the leg), we cut off all meshes at a height of 10 cm for evaluation.
We evaluate the quality of our 3D optimization method by comparing the output mesh to a ground truth scan from our dataset. We select a sample of 10,000 points from each mesh, sampling uniformly over the surface area. For each point, we find the nearest neighbor (NN) on the surface of the other mesh, and capture the difference in Euclidean distance (chamfer error), and the angular difference between the surface normals at the two points (normal error).

We report the statistics of the 20,000 samples across the two meshes. We compare our methods against photogrammetry pipeline COLMAP \cite{schonberger2016structure, schonberger2016pixelwise}, and surface normal optimization method FOUND \cite{boyne2024found}.

\paragraph{Speedtesting.} We evaluate the speed and requirements of all methods. All methods were evaluated on the same machine, with an NVIDIA Quadro P2200 GPU.

\paragraph{Number of views.} We vary the input views available to each method to investigate how this affects reconstruction quality. Where views are reduced, we sample views evenly in the left-right direction.
\section{Results}

\begin{table}[]
    \centering
    
    \begingroup
    \begin{footnotesize}
    \setlength{\tabcolsep}{4pt} 
    
\begin{tabular}{c|*3{c}|*3{c}}
    \toprule
     \multirow{2}{*}{Method} & \multicolumn{3}{c|}{NN chamfer error (mm) $\downarrow$} & \multicolumn{3}{c}{NN normal error ($^{\circ}$) $\downarrow$} \\
     & Mean & Median & RMSE & Mean & Median & RMSE \\ \midrule

    \colmap & \B 1.8 & \B 0.9 & 3.1 & 21.2 & 13.5 & 30.4 \\
    \found & 2.6 & 2.2 & 3.3 & \B 13.4 & \B 9.9 & 19.5 \\
    \ourSfM & 2.0 & 1.5 & \B 2.7 & 14.1 & 11.1 & \B 18.0 \\
    \ourOptim & 2.1 & 1.8 & \B 2.7 & 13.5 & 10.2 & 18.8 \\

     \bottomrule
\end{tabular}

    \end{footnotesize}
    \endgroup
    \caption{\textbf{3D reconstruction results.} Our methods yield the lowest RMSE chamfer error. While COLMAP still leads on mean and median chamfer error, it performs substantially worse on surface normal error, showing a substandard surface reconstruction. Our methods perform comparably to FOUND on surface normal reconstruction, despite \ourSfM using no 3D prior in the Poisson reconstruction stage, and \ourOptim not using surface normals during optimization.}
    \label{tab:3d_fit_results}
\end{table}

\begin{figure}
    \centering
    \includegraphics[width=\linewidth]{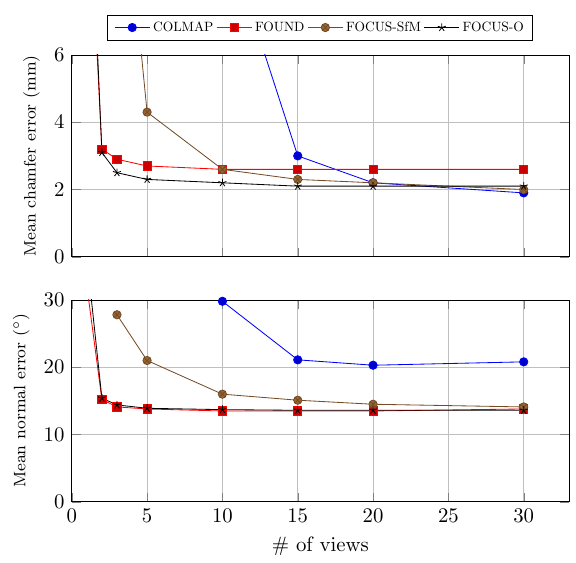}
    \caption{\textbf{Reconstruction quality as number of views varies.} \ourOptim shows the best performance under a few-view setting, retaining its accuracy with as few as 3 views. \ourSfM performs substantially better than COLMAP, showing accurate reconstruction performance with 10 views, compared to COLMAP requiring more than 15.}
    \label{fig:num_views}
\end{figure}

\begin{figure*}
    \centering
    \begin{tabular}{c|c}
        \includegraphics[height=0.325\linewidth]{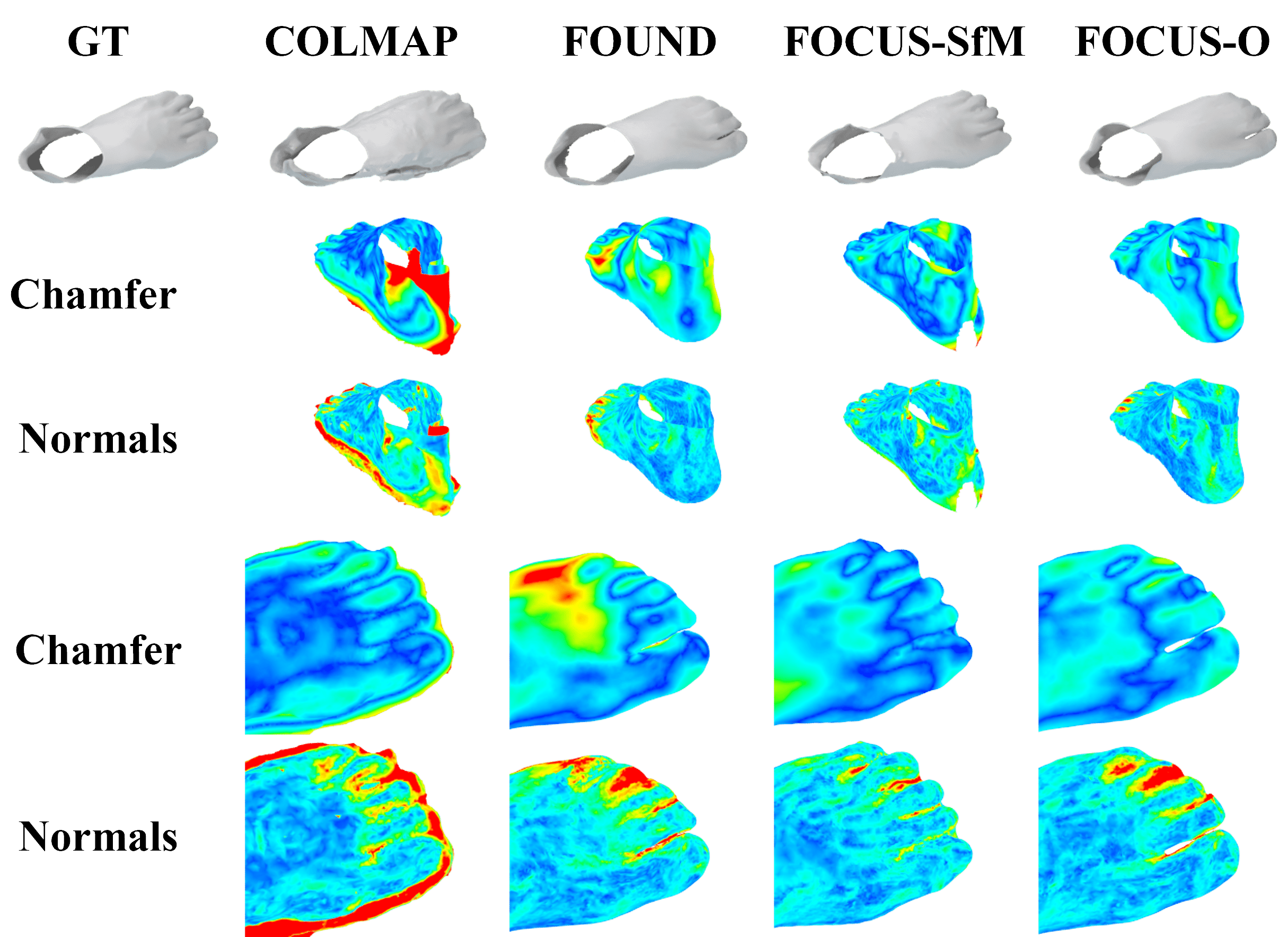} &
        \includegraphics[height=0.325\linewidth]{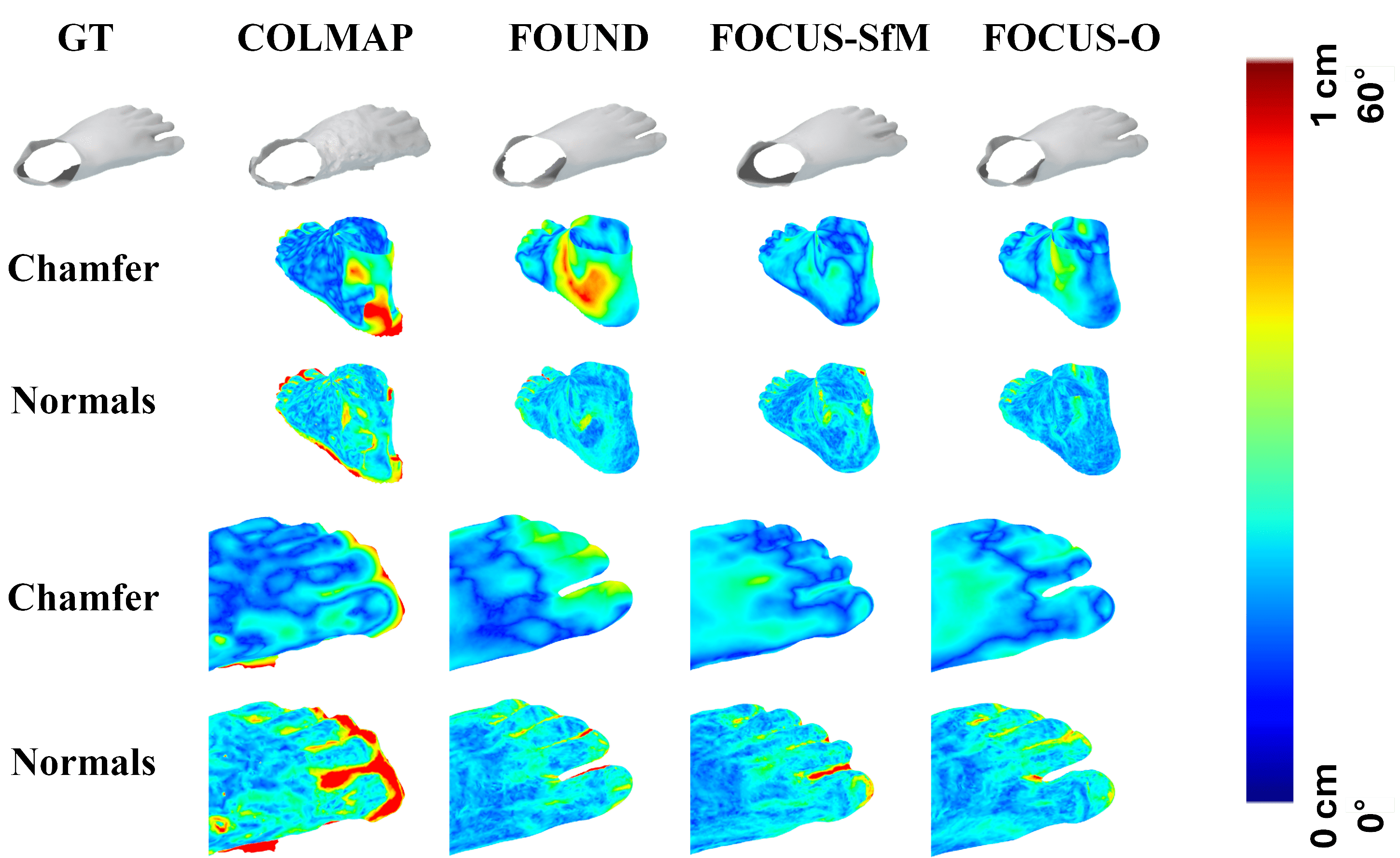}
    \end{tabular}
    \caption{\textbf{Qualitative reconstruction results.} The reconstruction quality is compared across two scans in the Foot3D dataset, comparing COLMAP, FOUND, \ourSfM and \ourOptim. COLMAP is prone to noise around the foot boundaries, and captures less surface detail in particular around the toes and heel.
    Further qualitative comparisons can be found in the supplementary material.}
    \label{fig:3d_fits_qual}
\end{figure*}


\paragraph{Reconstruction accuracy. } We show in Table \ref{tab:3d_fit_results} the accuracy of our two reconstruction methods, compared to FOUND and COLMAP, when all views are available (25-40) for each scan.

When all views are available, COLMAP performs the best for mean and median chamfer distance. However, our methods perform better on chamfer RMSE as COLMAP is prone to noise in reconstruction. Furthermore, our methods far outperform COLMAP with regards to surface normal quality, and in fact perform comparably with FOUND, which relies entirely on surface normals for fitting. \ourOptim uses no surface normals in the reconstruction process - only requiring TOC correspondences.

Detailed qualitative results can be seen in Figure \ref{fig:3d_fits_qual}. These show that COLMAP fits often exhibit noise around edges, which is problematic for taking measurements. Our methods are also superior at capturing the surface around the toes - the separation of the toes is handled better by both \ourSfM and \ourOptim.

\ourOptim tends to have fewer large sources of error both for normal and chamfer, providing a more reliable result. This approach, along with FOUND, produces watertight, parameterized meshes as output.

\paragraph{Number of views. } We show the performance under a varying number of views in Figure 
\ref{fig:num_views}. The strong 3D priors, and use of uncertainty, in FOUND and FOCUS-O ensure that they retain their reconstruction accuracy with as few as 3 views.
COLMAP has the highest view requirement, showing a significant reduction in accuracy (and some outright failures in reconstruction) for fewer than 15 views. While reconstructing using some similar principles to COLMAP, \ourSfM is able to take advantage of the rich information provided by TOCs to produce accurate reconstructions with as few as 10 views.

\begin{table}
    \centering
    \begingroup
    \small
    \newcommand{\cmark}{\ding{51}}%
\newcommand{\xmark}{\ding{55}}%


\begin{tabular}{c|cc|cc}
    \toprule
    \multirow{2}{*}{Method} & \multicolumn{2}{c|}{Inference time (s)} & \multirow{2}{*}{GPU?} & Differentiable\\
    & 10-view & all-view & & rendering? \\
    \midrule
    \colmap & - & 465 & \cmark & \xmark \\
    \found & 260 & 260 & \cmark & \cmark \\
    \ourSfM & \B 22 & \B 89 & \xmark & \xmark\\
    \ourOptim & 40 & 100 & \cmark & \xmark \\

    \bottomrule
\end{tabular}



    \endgroup
    \caption{\textbf{Speed and requirements.} \ourOptim and \ourSfM are substantially faster than existing methods. Furthermore, neither require differentiable rendering, and \ourSfM does not require a GPU.}
    \label{tab:speed_and_requirements}
\end{table}

\paragraph{Implementation.} Table \ref{tab:speed_and_requirements} compares the inference times and computational requirements of all methods. Both of our methods are substantially faster than COLMAP and FOUND. Compared to FOUND, neither require differentiable rendering, drastically reducing the memory requirements, and making them easier to implement on devices that would not support it. Further, \ourSfM is a CPU based method (except for the initial TOC predictions), so is possible to run on devices without GPU optimization support, including mobile devices.

\begin{table}
    \centering
    \begingroup
    \footnotesize
    \centering
    
\begin{tabulary}{0.9\linewidth}{cCC}
    \toprule
      & NN mean chamfer error (mm) $\downarrow$ & NN mean normal error ($^{\circ}$) $\downarrow$ \\ \midrule
    \ourSfM & \B 2.0 & \B 14.0 \\
    w/o subpixel matching & 2.4 & 15.4 \\
    w/o normal aggregation & 2.5 & 107.5 \\
 \bottomrule
\end{tabulary}

    \endgroup
    \caption{\textbf{\ourSfM ablation study.} Both the subpixel matching and normal aggregation steps of \ourSfM are crucial for its quantitative reconstruction performance. The drastic increase in normal error without normal aggregation is partly due to the reconstruction algorithm occasionally inverting the surface normals.}
    \label{tab:ablation-sfm}
\end{table}

\begin{table}
    \centering
    \begingroup
    \footnotesize
    \centering
    \begin{tabular}{c|cc|cc}
      \toprule
       & \multicolumn{2}{c|}{\scriptsize {NN Chamfer error (mm) $\downarrow$}} & \multicolumn{2}{c}{\scriptsize NN Normal error ($^{\circ}$) $\downarrow$}\\
       & 3 view & 20 view & 3 view & 20 view \\
       \midrule
    \ourOptim & \B 2.5 &  \B 2.1 &  \B 14.4 & \B 13.5 \\
    w/o uncertainty &  2.7 &  2.2 &  15.0 &  13.9 \\

      \bottomrule
\end{tabular}
    \endgroup
    \caption{\textbf{\ourOptim ablation study.} The use of TOC uncertainty in our optimization process improves all reconstruction metrics, both for a low and high view count.}
    \label{tab:ablation-optim}
\end{table}

\paragraph{Ablation study.} We show results of our ablation for both \ourSfM and \ourOptim in Tables \ref{tab:ablation-sfm} and \ref{tab:ablation-optim} respectively.

For \ourSfM, aggregating the predicted surface normals is critical to capturing geometry in certain areas of the foot, especially around the toes, as naive Poisson reconstruction is not capable of estimating that detail. Furthermore, Poisson reconstruction may generate a plausible mesh with inverted face normals, hence the high surface normal error.
Our subpixel matching is also critical for capturing some of the finer details on the foot surface.

For \ourOptim, the use of uncertainty in weighting the various TOC samples used for fitting the FIND model provides a boost to fitting accuracy.
\section{Conclusion}

We have shown that the TOC representation provides a strong signal for predictors to identify dense correspondences on in-the-wild images, and that it is possible to train these models using entirely synthetic data. Our synthetic dataset, \ourSynth, provides substantial improvements on SynFoot, including more background and lighting variation, and adding articulation to the 3D models used for rendering.

We have shown that TOCs can be combined with uncertainty and surface normal predictions to recover high quality foot reconstructions from multiple views.

The first of our reconstruction methods, \ourSfM, effectively identifies subpixel TOC matches across multiple images, and triangulates these to provide a high quality reconstruction with no enforcement of a 3D prior on the output foot shape. This method requires fewer views than COLMAP, and can operate entirely on a CPU.

Our second method, \ourOptim, is capable of directly optimizing the parameterized FIND model to match the TOC predictions across images. \ourOptim effectively uses TOC uncertainty, and provides a superior reconstruction to FOUND, while matching the low view requirement, and running substantially faster.

Each method has its own advantages. \ourSfM may be suitable for applications where speed is desired, no GPU is available, and only the visible portion of the surface is of interest to downstream tasks. In contrast, \ourOptim may be more desirable when a fully parameterized and watertight mesh is required.
\section{Limitations and future work}

Our methods currently rely on calibrated cameras for reconstruction. However, the dense correspondences could be used to recover the relative camera poses, up to a scale ambiguity. We did not explore this as part of this paper, as our evaluations are against Foot3D, requiring accurate camera scale and pose. This approach is implemented in the code released alongside this paper, and can be run on an arbitrary set of uncalibrated images.

\ourOptim does not leverage the image surface normal predictions, as there is no simple mapping from TOC to surface normals, and so would require differentiable rendering. Modifying the approach to use surface normals could improve accuracy, at the cost of computation time. Similarly, \ourSfM does not use the predicted TOC uncertainty, as we did not find it to improve the reconstruction. Future research could identify an approach to use this signal to improve \ourSfM's reconstruction.\\
\section{Acknowledgments}

\sloppy
The authors acknowledge the collaboration and financial support of Trya Srl.
\fussy
{
    \small
    \bibliographystyle{ieeenat_fullname}
    \bibliography{main}
}
\clearpage
\setcounter{page}{1}

\maketitlesupplementary

\section{Further examples}

\begin{figure}[!h]
    \centering
    \begin{minipage}{.8\linewidth}
      \centering

    \begin{tabular}{*{4}{@{}Q{0.2375\linewidth}}@{}}
    (a) & (b) &  (c) & (d)
    \end{tabular}
      
      \includegraphics[width=0.95\linewidth]{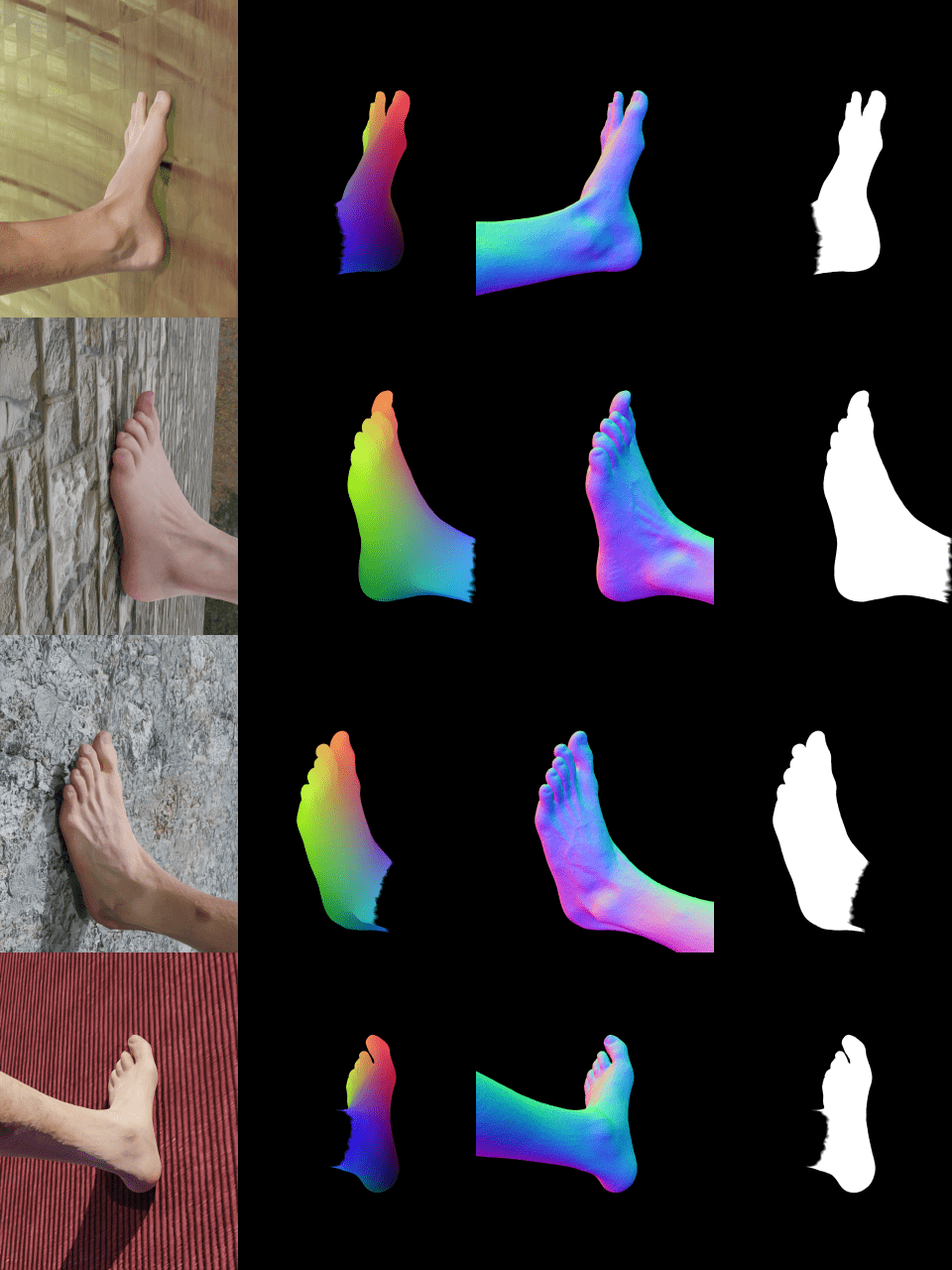}
    \end{minipage}%

    \vspace{10pt}
    
    \begin{minipage}{.8\linewidth}
      \centering

      
      \includegraphics[width=0.95\linewidth]{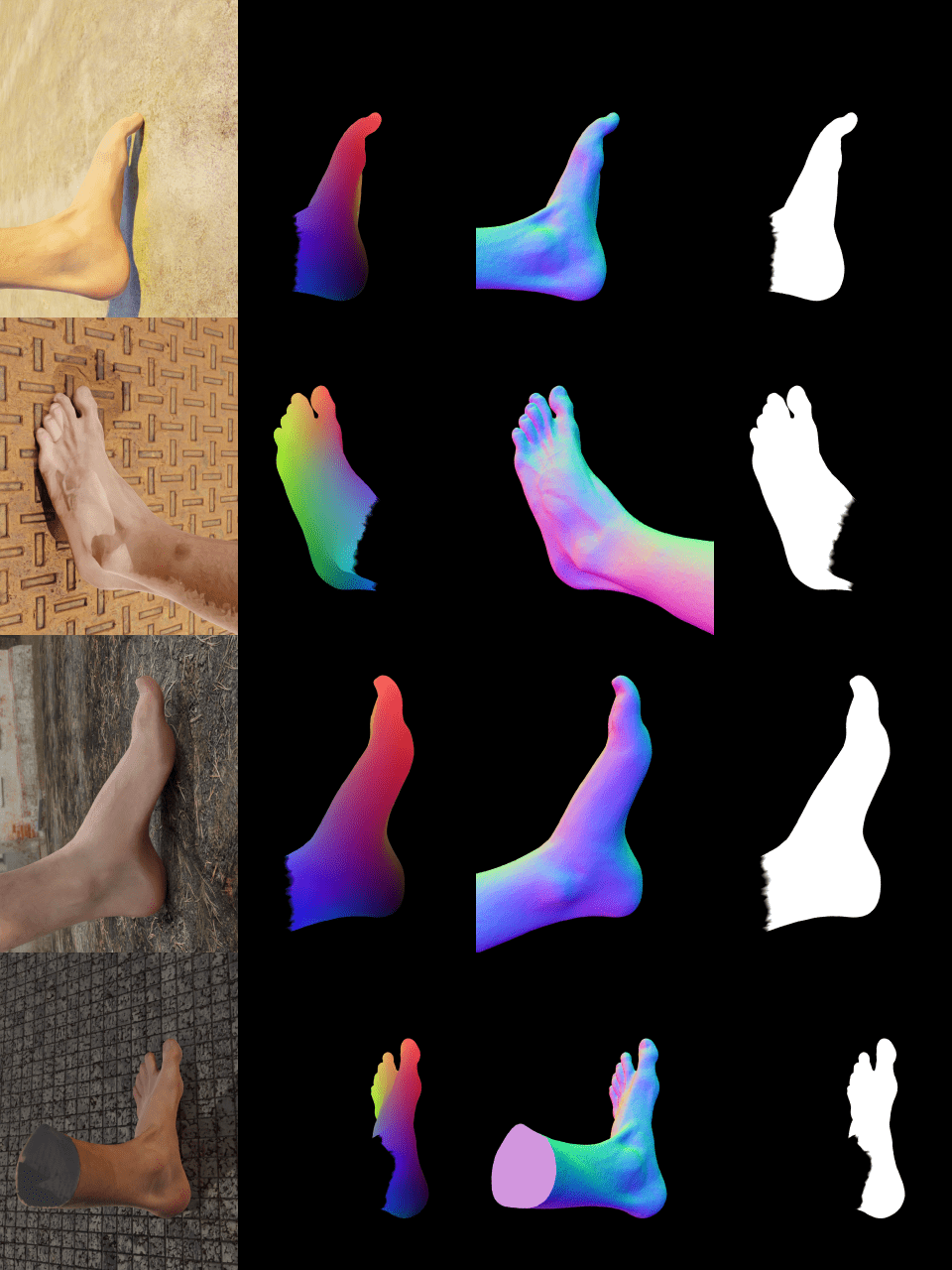}
    \end{minipage}
    
    \caption{\textbf{SynFoot2 examples.} Further examples of SynFoot2, showing (a) RGB, (b) TOC, (c) surface normals, and (d) segmentation masks.}
    \label{fig:synth-supp}
    
\end{figure}

\paragraph{Synthetic dataset.} In Figure \ref{fig:synth-supp}, we show additional samples of our synthetic dataset.

\begin{figure*}
    \centering
    \begin{minipage}{.4\linewidth}
      \centering

    \begin{tabular}{*{4}{@{}Q{0.2375\linewidth}}@{}}
    (a) & (b) &  (c) & (d)
    \end{tabular}
      
      \includegraphics[width=0.95\linewidth]{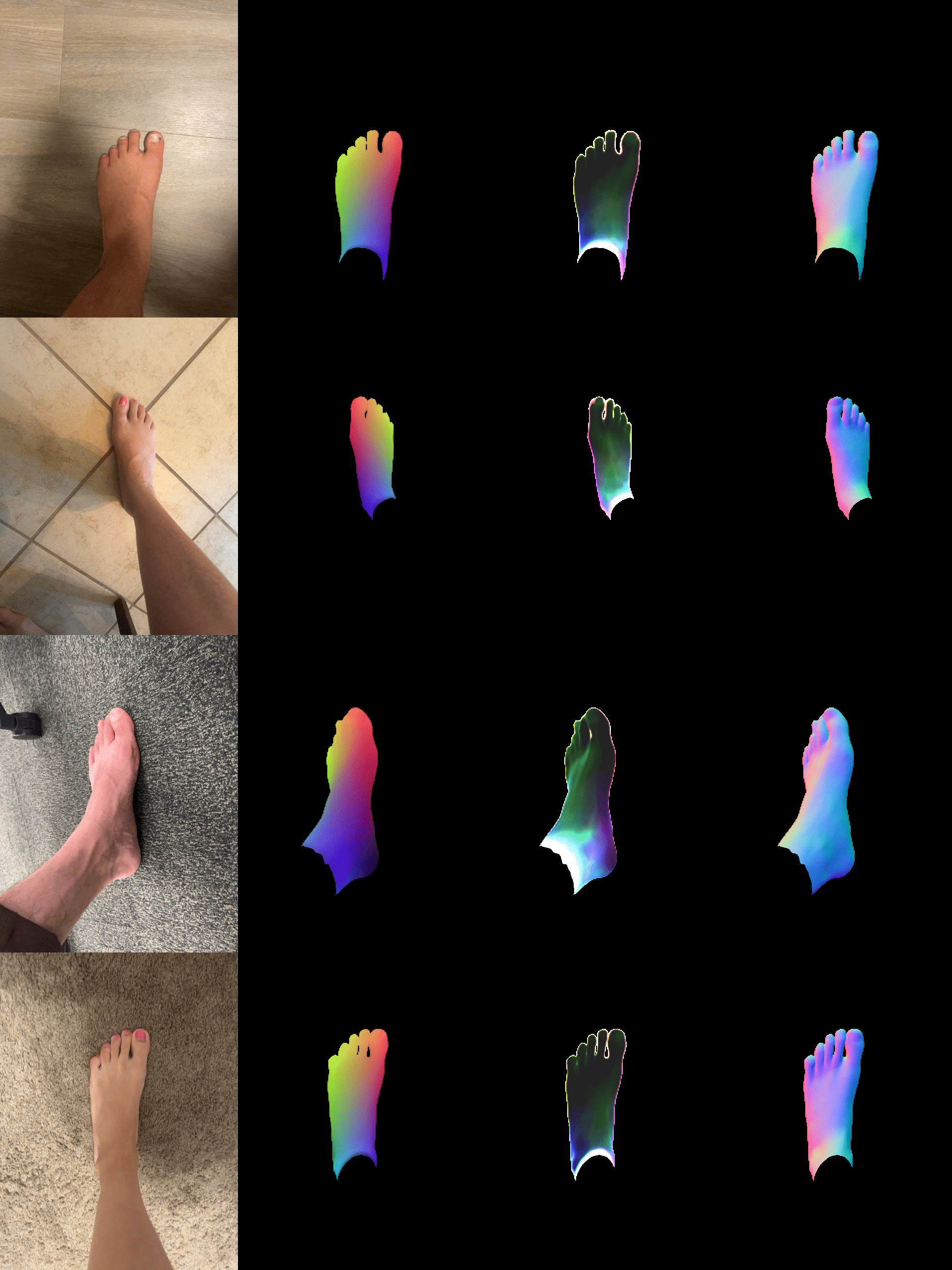}
    \end{minipage}%
    \begin{minipage}{.4\linewidth}
      \centering

    \begin{tabular}{*{4}{@{}Q{0.2375\linewidth}}@{}}
    (a) & (b) &  (c) & (d)
    \end{tabular}
      
      \includegraphics[width=0.95\linewidth]{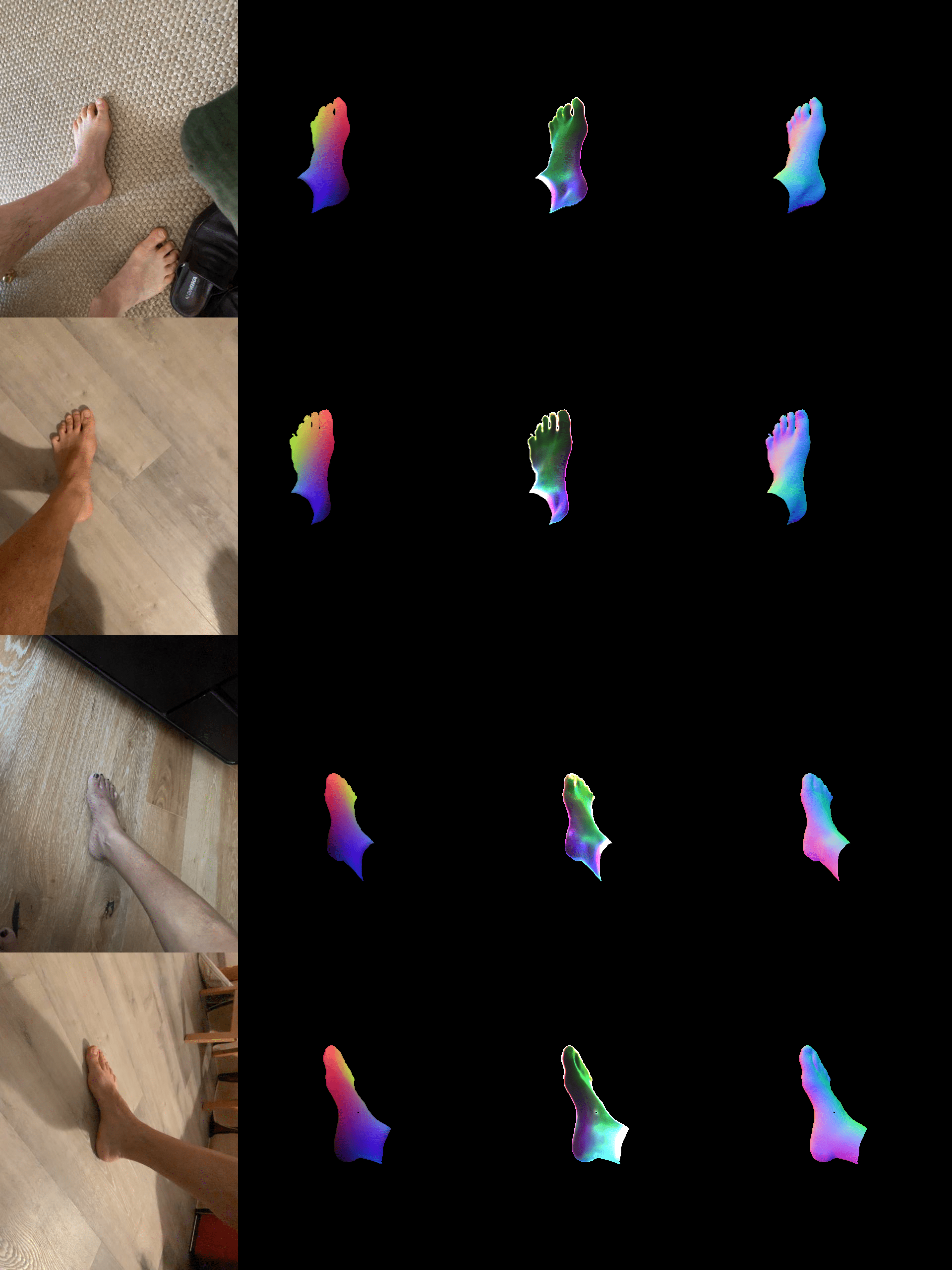}
    \end{minipage}%
    \begin{minipage}{.15\textwidth}
      \centering
      \includegraphics[width=0.7\linewidth]{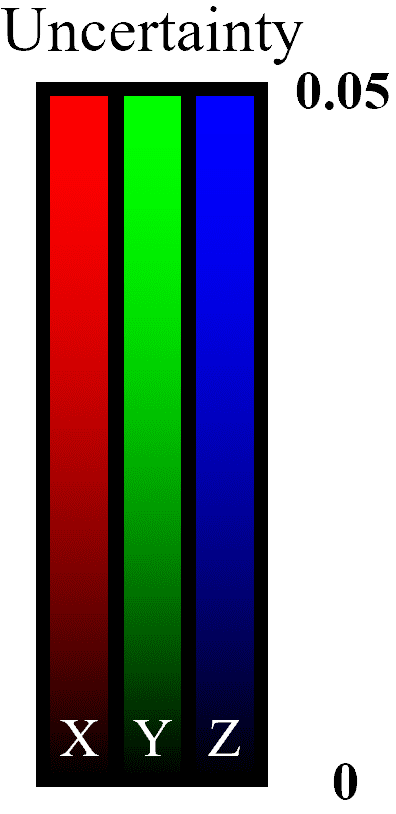}
    \end{minipage}
    
    \caption{\textbf{In-the-wild predictions.} Further examples of in-the-wild predictions of our TOC predictor, showing showing (a) RGB input, (b) TOC $\toc$, (c) TOC uncertainty $\tocstd$, and (d) surface normals. We show some more challenging examples in the right figure, even showing robustness to a secondary foot in the image.}
    \label{fig:itw-supp}
    
\end{figure*}

\paragraph{In-the-wild predictions.} In Figure \ref{fig:itw-supp}, we show further qualitative predictions of our TOC and normal predictions on in-the-wild images.

\begin{figure*}
\centering
\newcommand\qualfig[1]{\includegraphics[width=0.45\linewidth]{images/supp/qual/#1_comp.png}}
\begin{tabular}{ccc}
\qualfig{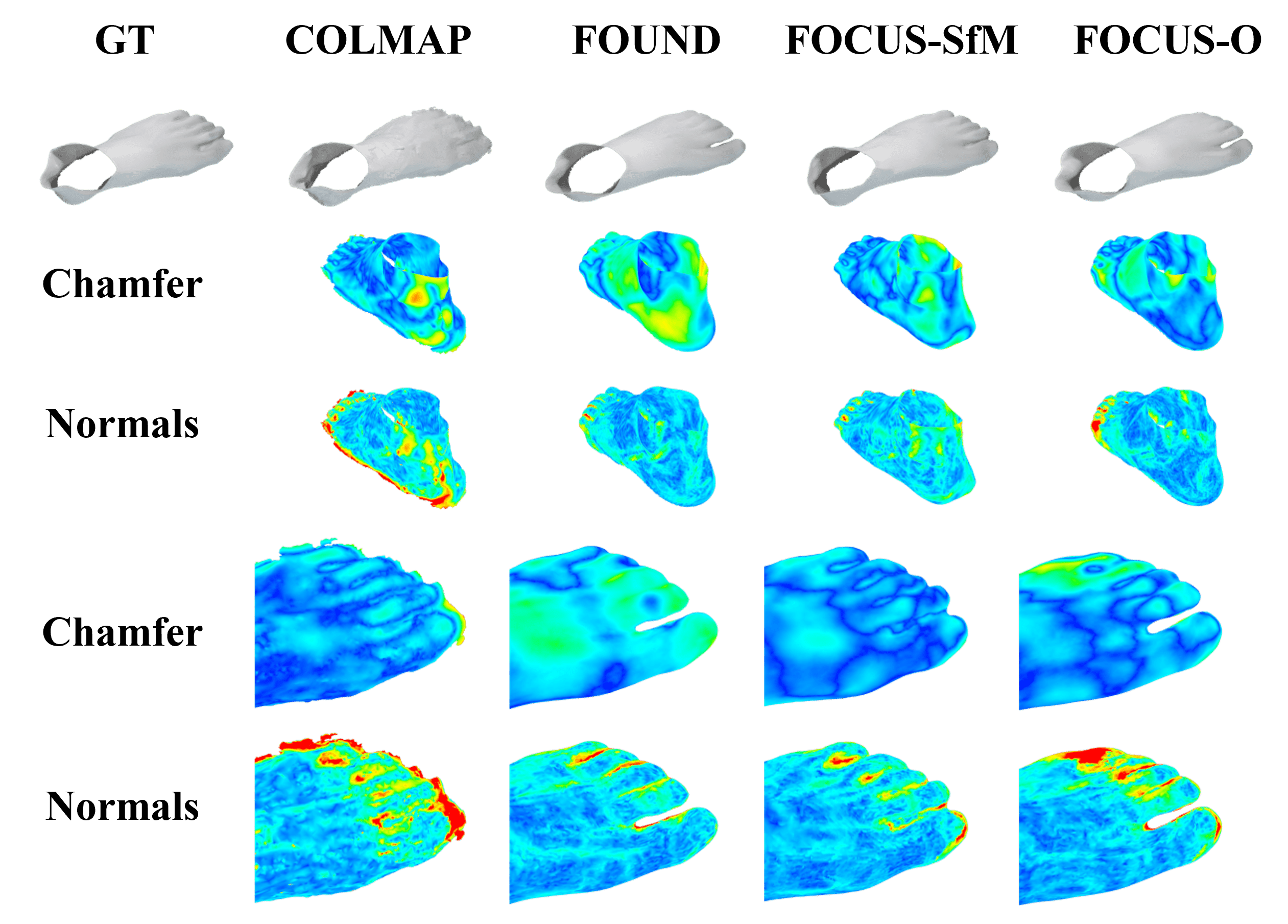} & \qualfig{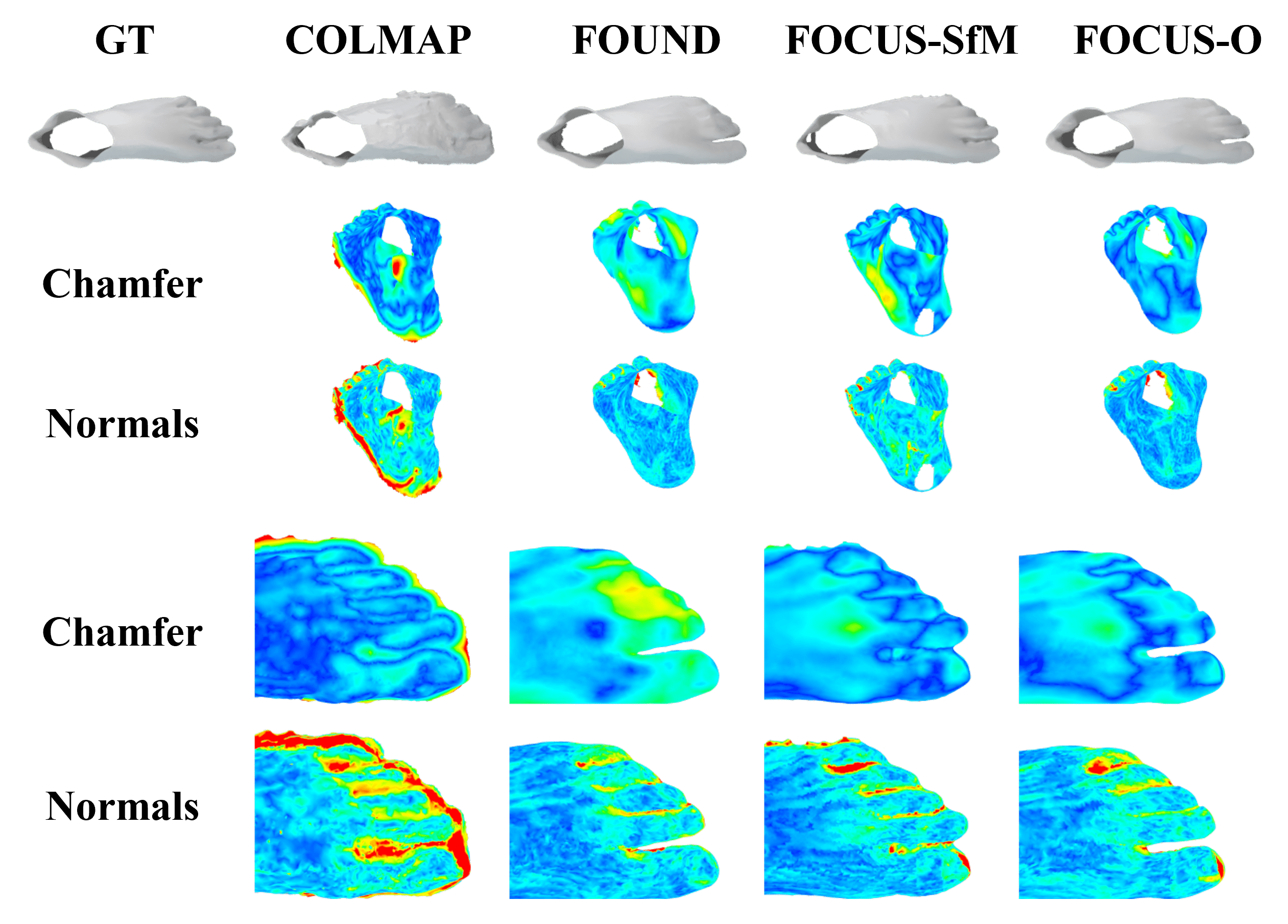} & \multirow{2}{*}[\dimexpr 1.5cm]{\includegraphics[width=0.05\linewidth]{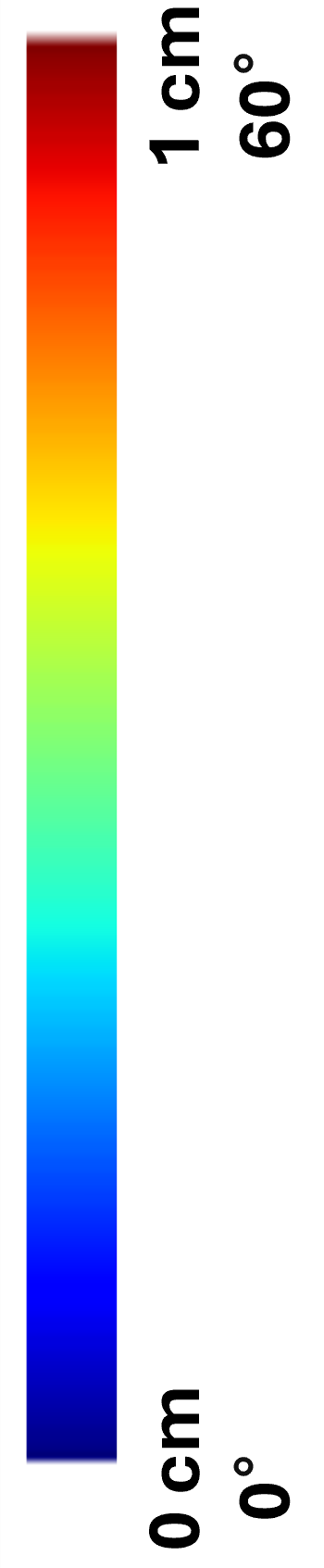}}\\[10pt]
\qualfig{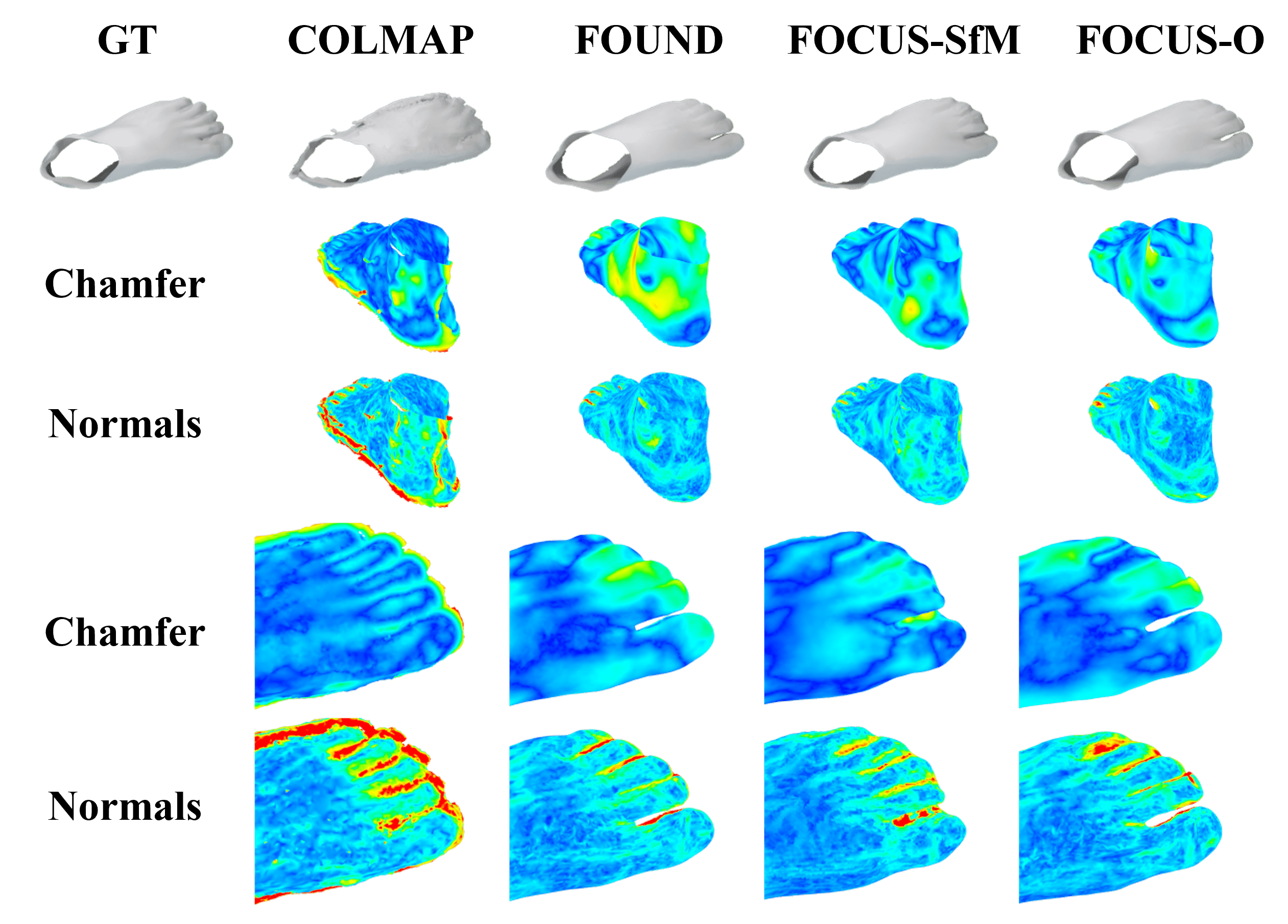} & \qualfig{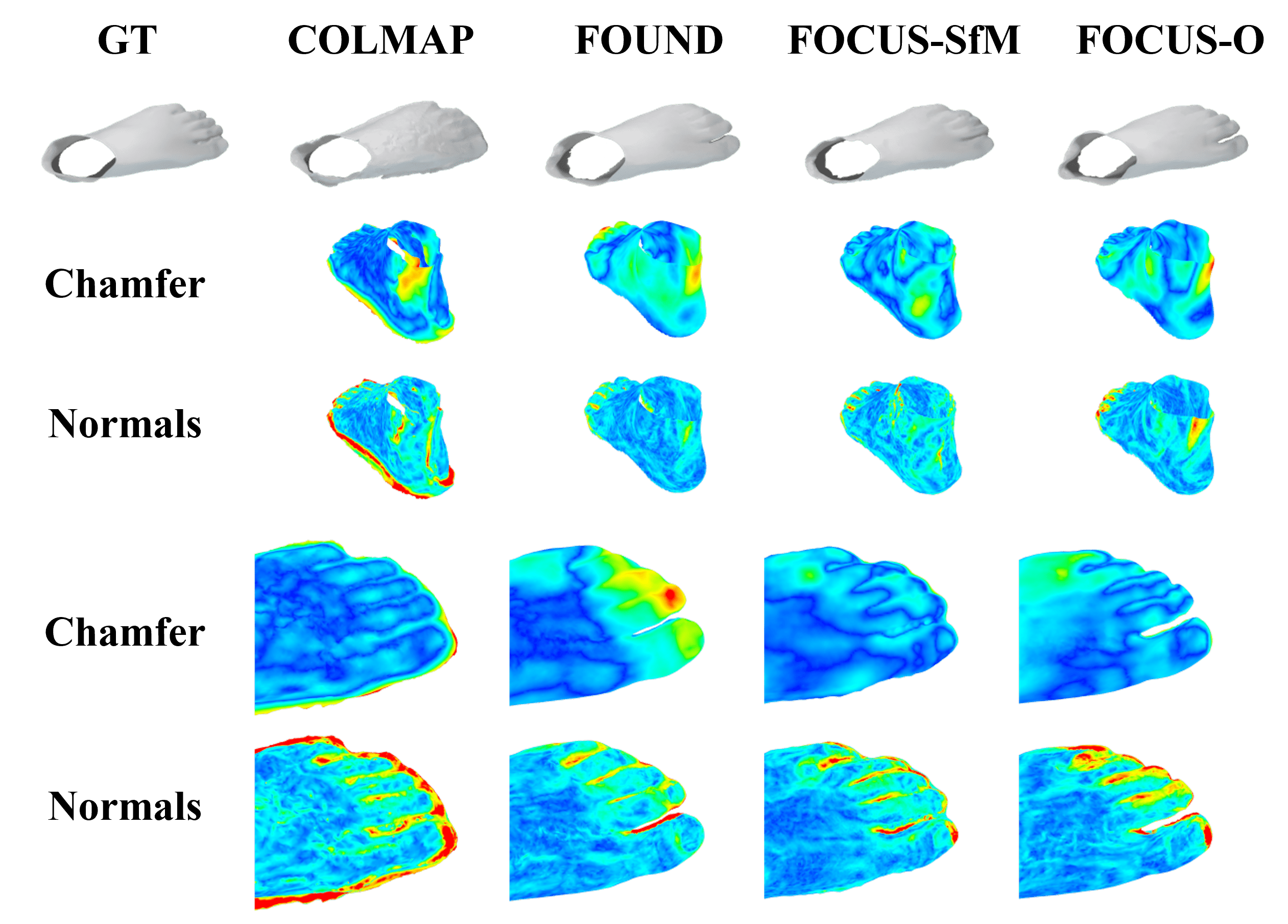} \\

\end{tabular}
\caption{\textbf{Qualitative reconstruction results.} The reconstruction quality is compared across four further scans in the Foot3D dataset, comparing COLMAP, FOUND, \ourSfM and \ourOptim.}
\label{fig:qual-supp}
\end{figure*}

\paragraph{Reconstruction.} In Figure \ref{fig:qual-supp}, we show further qualitative reconstruction comparison to existing methods FOUND \cite{boyne2024found} and COLMAP \cite{schonberger2016structure, schonberger2016pixelwise}.

\section{Method hyperparameters}

We define the hyperparameters used for \ourSfM in Table \ref{fig:sfm-hyper}.

\begin{table*}
    \centering
    
\begin{tabular}{Q{0.6\linewidth}cc}
    \toprule
    Description & Value & Unit \\\midrule
    Number of samples taken per image & 3000 & - \\
    Maximum $\ell 2$ distance to consider a correspondence a match & 0.002 & - \\
    Upsampling factor for subpixel matching & 8 & - \\
    Reprojection error above which to filter triangulated points&  3 & pixels \\
    Crop the mesh to this padding around reconstructed point cloud & 1 & mm \\
    Foot mesh height interval & [0, 150] & mm \\
    Poisson reconstruction - depth & 8 & - \\
    Poisson reconstruction - iterations & 8 & - \\
    \bottomrule
\end{tabular}
    \caption{Hyperparameters chosen for \ourSfM.}
    \label{fig:sfm-hyper}
\end{table*}

\end{document}